%% file: neurips_2026.tex
\documentclass{article}

\PassOptionsToPackage{numbers,compress}{natbib}
\usepackage[preprint]{neurips_2026}
\usepackage[utf8]{inputenc}
\usepackage[T1]{fontenc}
\usepackage{hyperref}
\usepackage{url}
\usepackage{booktabs}
\usepackage{amsfonts}
\usepackage{amsmath}
\usepackage{amssymb}
\usepackage{microtype}
\usepackage{xcolor}
\usepackage{graphicx}
\usepackage{multirow}
\usepackage{makecell}
\usepackage{tabularx}
\usepackage{longtable}
\usepackage{array}
\usepackage{colortbl}
\usepackage{caption}
\usepackage[caption=false,font=normalsize,labelfont=sf,textfont=sf]{subfig}
\usepackage{algorithm}
\usepackage{algorithmic}
\usepackage{pifont}
\newcommand{\cmark}{\ding{51}} 
\newcommand{\xmark}{\ding{55}} 

\definecolor{LightCyan}{rgb}{0.88,1,1}
\definecolor{HighLight}{rgb}{0.96,0.92,0.96}

\title{SteelDefectX: A Multi-Form Vision-Language Dataset and Benchmark for Steel Surface Defect Analysis}

\author{{Shuxian Zhao$^1$ \quad Jie Gui$^{1,2,}$\thanks{Corresponding author.} \quad Baosheng Yu$^3$ \quad Dacheng Tao$^3$} \\
	$^1$Southeast University \quad $^2$Purple Mountain Laboratories  \quad $^3$Nanyang Technological University  \\
}

\begin{document}
	
	\maketitle
	
	\begin{abstract}
	Steel surface defect analysis is critical for industrial quality control, yet existing benchmarks rely primarily on label-only annotations, limiting fine-grained semantic understanding and systematic evaluation of vision-language models.
	To address this gap, we introduce SteelDefectX, a vision-language dataset with multi-form textual annotations for steel surface defect analysis, comprising 7,778 images across 25 defect categories. At the class level, the dataset provides defect names, representative visual attributes, and industrial causes. At the sample level, each image is annotated with three forms of textual representations: (1) free-form natural language descriptions, (2) structured attribute annotations, and (3) template-based sentences. These annotations provide flexible textual supervision with varying levels of expressiveness and controllability.
	We further establish a comprehensive benchmark covering vision-language classification, segmentation, and cross-dataset transfer, along with additional evaluations such as retrieval and text-guided localization.
	Experimental results reveal a trade-off between structure and flexibility in textual representations. Structured attributes provide more stable semantic alignment, while natural language descriptions improve transferability and fine-grained spatial grounding. These findings highlight the critical role of textual design in industrial vision-language learning.
	SteelDefectX provides a new benchmark for studying semantic alignment and generalization in industrial vision-language learning. The code and dataset are available at \url{https://github.com/Zhaosxian/SteelDefectX}.
	\end{abstract}
	
	\input{steeldefectx_body.tex}

	\bibliographystyle{unsrtnat}
	\bibliography{main}
	
	\appendix
	\newpage
    \input{Appendix.tex}
	
\end{document}

%% file: steeldefectx_body.tex
\section{Introduction}
\label{sec:intro}
Steel surface defect analysis is critical for ensuring the quality and reliability of industrial steel products~\cite{Wen2022Steel, Ma2024Surface}. Despite the success of deep visual models in industrial inspection~\cite{he2016deep, Vit2021an}, existing steel surface defect detection benchmarks remain largely limited to classification, detection, or segmentation tasks with coarse-grained annotations~\cite{zheng2025legendre,ma2025ela,huang2025multi,luo2018generalized, liu2025global}. These datasets typically provide only class labels or numerical annotations, which are insufficient for capturing the rich semantic properties of defects, such as their visual characteristics, spatial distribution, and potential causes. As a result, current benchmarks are not well-suited for evaluating emerging vision-language models~\cite{radford2021learning, li2022blip, liu2023visual, laurenccon2024matters}, which rely on expressive and structured textual supervision.

Recent attempts to extend industrial defect analysis into the vision-language paradigm often rely on simple template-based descriptions derived from class labels~\cite{Jeong2023WinCLIP, Lei2025Improving, ma2025aa}. However, such templates fail to capture the intrinsic complexity of steel surface defects. In practice, defects exhibit high variability in appearance due to differences in materials, manufacturing processes, and environmental conditions. Even within the same defect category, visual patterns can vary significantly, making it difficult for models to generalize based solely on class-level descriptions. This highlights a fundamental limitation of existing benchmarks: they lack fine-grained, sample-specific semantic annotations that reflect how defects are described and analyzed in real industrial settings.

\begin{figure}[t]
	\centering
	\includegraphics[width=\linewidth]{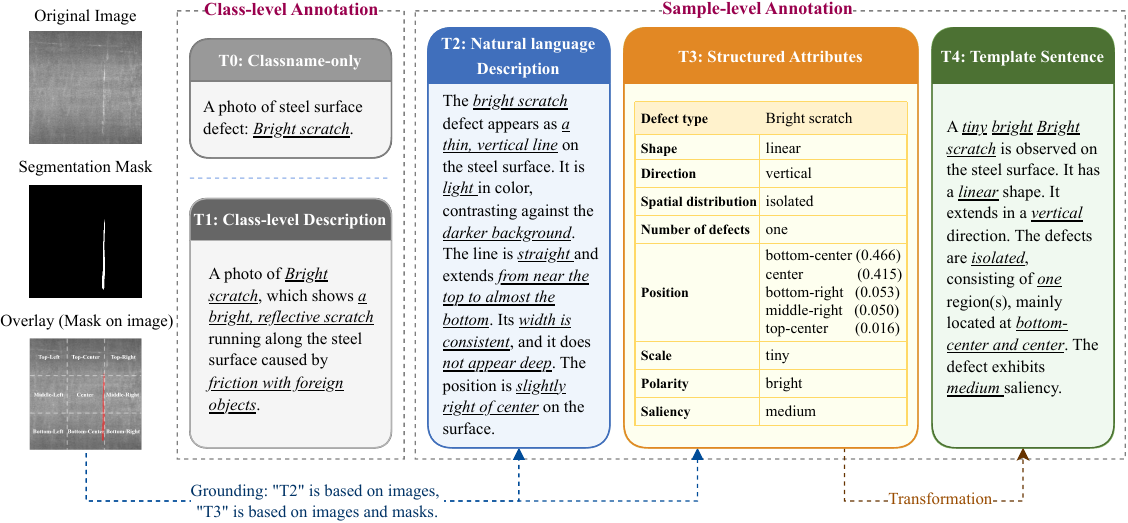}
	\caption{
		Illustration of multi-form textual annotations in SteelDefectX. Given a defect image and its segmentation mask, we present text forms from class-level descriptions (T1) to sample-level annotations, including natural language descriptions (T2), structured attributes (T3), and template-based sentences (T4). These forms vary in expressiveness and structure while sharing consistent attribute-based representations, enabling diverse forms of supervision for vision-language learning.
	}
	\vspace{-0.1cm}
	\label{fig:fig1}
\end{figure}

To address this gap, we introduce \textbf{SteelDefectX}, a multi-form vision-language dataset and benchmark for steel surface defect analysis. SteelDefectX is constructed by unifying four widely used public datasets (NEU~\cite{song2013noise}, GC10~\cite{lv2020deep}, X-SDD~\cite{feng2021x}, and S3D~\cite{cui2024efficient}) into a consistent taxonomy of 25 defect categories, comprising 7,778 images. More importantly, our dataset augments each image with multi-level textual annotations that bridge low-level visual observations and high-level semantic reasoning (see Fig.~\ref{fig:fig1}). At the class level (T1), each defect category is described using its name, representative visual attributes, and potential industrial causes. At the sample level, we design three complementary forms of textual annotations: (1) free-form natural language descriptions (T2), (2) structured attribute annotations (T3), and (3) template-based sentences (T4). ``T2'' provides more flexible semantic representations, ``T3'' supports controllable and interpretable supervision, and ``T4'' offers a standardized intermediate format. These forms vary in expressiveness and structure while sharing a consistent attribute-based design, enabling flexible supervision for different vision-language tasks. These annotations also enable systematic analysis of how different textual formats influence vision-language learning. Pixel-level segmentation masks are also provided.

The sample-level annotations are constructed through tailored pipelines.``T2'' is generated through a multi-stage pipeline combining large language models (LLMs)~\cite{hurst2024gpt}, semantic-similarity filtering, dimension coverage evaluation, and manual refinement to ensure annotation consistency and quality.   ``T3'' is extracted through a hybrid process that combines mask-based computation, image statistics, and constrained semantic inference. Attributes such as position, scale, polarity, saliency, and the number of defects are derived from segmentation masks and pixel-level measurements, while shape and direction are inferred from predefined candidate sets based on ``T2''. Based on ``T3'', ``T4'' converts structured attributes into standardized template sentences. This design allows SteelDefectX to support both expressive reasoning and controllable semantic modeling.

Building on SteelDefectX, we establish a unified benchmark for evaluating vision-language learning in industrial defect analysis. The benchmark covers vision-language classification~\cite{radford2021learning, Cherti2023Reproducible, sun2023eva, Zhang2024LongCLIP, xie2025fgclip}, vision-language segmentation, cross-dataset transfer~\cite{Wang2024SkyScript, liATPrompt2025}, and further analyses on structured attributes, vision-language retrieval~\cite{meng2025evdclip}, text-guided segmentation, and image-text similarity. These tasks allow us to systematically compare different text forms and examine how textual design affects classification, cross-modal alignment, dense localization, and transferability. Experimental results show that structured attributes tend to provide stable and effective supervision for classification and retrieval, while free-form natural language descriptions often exhibit stronger transferability to unseen datasets and better fine-grained grounding in text-guided segmentation. These findings suggest that different textual representations offer complementary benefits and should be selected according to the target task. The main contributions of this paper are summarized as follows:

\begin{itemize}
	\item We propose SteelDefectX, a steel surface defect dataset with multi-form vision-language annotations. It includes 7,778 images across 25 categories and provides class-level descriptions, sample-level textual annotations, and pixel-level segmentation masks.
	
	\item We design a multi-form sample-level annotation framework comprising natural language descriptions (T2), structured attributes (T3), and template-based sentences (T4). This design supports both expressive semantic descriptions and controllable textual representations for industrial vision-language learning.
	
	\item We establish a comprehensive benchmark covering multiple vision-language tasks and settings. We analyze how different textual representations affect model performance, offering new insights into vision-language learning for industrial inspection.
\end{itemize}

\section{Related Work}
\label{sec: related work}
Despite recent advances in industrial anomaly detection and multimodal learning, existing datasets remain inadequate for fine-grained and controllable vision-language analysis. As shown in Table~\ref{tab:dataset_comparison}, early industrial defect datasets, such as NEU~\cite{song2013noise}, GC10~\cite{lv2020deep}, X-SDD~\cite{feng2021x}, and S3D~\cite{cui2024efficient}, mainly support visual recognition and provide only class labels, lacking semantic textual supervision. Although anomaly detection benchmarks, including MVTec-AD~\cite{MVTecAD2021} and VisA~\cite{zou2022spot}, introduce pixel-level annotations, they remain limited to vision-only settings and do not support language-guided understanding. Recent multimodal datasets attempt to address this limitation. MMAD~\cite{jiang2025mmad} provides question–answer pairs for multimodal reasoning, while MVTec-Caption~\cite{hu2024anomalyxfusion} and IMDD-1M~\cite{ni2026towards} offer image-text pairs with natural language descriptions. AnomVerse~\cite{jiang2026anomagic} includes an automatically generated template caption to support anomaly synthesis and detection. 

However, these datasets typically rely on structured, template-based textual representations. In particular, they do not explicitly model different levels of semantic abstraction, making it difficult to systematically study how textual design influences vision-language learning. SteelDefectX addresses these limitations by introducing a multi-form annotation framework that explicitly models different levels of semantic abstraction. It integrates class-level descriptions with sample-level annotations, including free-form natural language descriptions (T2), structured attributes (T3), and template-based sentences (T4). This design enables systematic analysis of how different textual representations influence vision-language alignment, generalization, and spatial grounding. 

\begin{table*}[t]
	\centering
	\small
	\caption{Comparison of SteelDefectX with representative industrial defect and multimodal datasets.}
	\label{tab:dataset_comparison}
	\resizebox{0.99\linewidth}{!}{
		\begin{tabular}{lccccc}
			\toprule
			\textbf{Dataset} & \textbf{Domain} & \textbf{\#Images}  & \textbf{Sample-level Text} & \textbf{Structured Attributes} & \textbf{Segmentation Mask}  \\
			\midrule
			NEU~\cite{song2013noise} & Steel & 1.8K & \xmark & \xmark & \xmark  \\
			GC10~\cite{lv2020deep} & Steel & 2.3K &  \xmark & \xmark & \xmark  \\
			X-SDD~\cite{feng2021x} & Steel & 1.3K  & \xmark & \xmark & \xmark \\
			S3D~\cite{cui2024efficient} & Steel & 0.8K  & \xmark & \xmark & \cmark \\
			\midrule
			MVTec-AD~\cite{MVTecAD2021} & Multi-object & 5.3K &  \xmark & \xmark & \cmark  \\
			VisA~\cite{zou2022spot} & Multi-object & 10.8K & \xmark & \xmark & \cmark \\
			\midrule
			MMAD~\cite{jiang2025mmad} & Industrial & 8.3K & Limited (Q\&A) & \xmark & \cmark  \\
			IMDD-1M~\cite{ni2026towards} & Industrial & 1M & \cmark & \xmark & \cmark  \\
			MVTec-Caption~\cite{hu2024anomalyxfusion} & Industrial & 2.2K & \cmark & \xmark & \cmark \\
			AnomVerse~\cite{jiang2026anomagic} & Industrial & 12.9K & \cmark & \xmark & \cmark  \\
			\midrule
			\textbf{SteelDefectX} & Steel & 7.8K & \cmark & \cmark & \cmark  \\
			\bottomrule
	\end{tabular}}
\end{table*}

\section{Dataset}
\label{sec:SteelDX}

\subsection{Data Collection}
\label{Subsec: Dataset_Construction}

The SteelDefectX dataset is constructed by integrating and reorganizing four publicly available steel surface defect datasets: NEU~\cite{song2013noise}, GC10~\cite{lv2020deep}, X-SDD~\cite{feng2021x}, and S3D~\cite{cui2024efficient}. To enhance coverage and representation, we also incorporate selected and processed samples from FSC-20~\cite{zhao2023fanet} and ESDIs-SOD~\cite{cui2024efficient}, as they are derived from similar data sources. The integration process involves three key steps. First, all images are standardized to a resolution of $256\times256$, and redundant or low-quality samples are filtered out to ensure data uniformity. Second, defect annotations are cross-verified and consolidated across sources to eliminate inconsistencies. Finally, visually and semantically similar subclasses are merged into a unified taxonomy of 25 categories, yielding a compact yet coherent label space that maintains consistency across datasets while preserving representational diversity. Details are provided in Appendix~\ref{sec:sup_a} and Appendix~\ref{subsec:class_visual}.

\subsection{Class-Level Annotation}

Motivated by the question, ``\textit{How can we effectively describe industrial defect images with natural language for multimodal understanding?}'', we design class-level descriptions for each of the 25 defect classes to provide global contextual semantics. Initial templates are manually crafted based on domain knowledge from steel manufacturing and subsequently refined using descriptions generated by CuPL~\cite{Pratt2023Platypus}. Each class-level description consists of three components: (1) \textit{the defect class name}; (2) \textit{representative visual attributes}, including shape, texture, and color; and (3) \textit{possible industrial causes}, such as rolling defects or material inclusions. These elements are combined into smooth, natural-language descriptions that capture shared semantic properties across samples within each category and provide consistent conceptual grounding for vision-language alignment.

\subsection{Sample-Level Annotation}
\label{subsec:sample_level}
While class-level annotations provide consistent category semantics, individual defect samples exhibit significant visual variation and require fine-grained descriptions. To address this, we provide multiple complementary forms of sample-level annotations, including free-form natural language descriptions (T2), structured attribute representations (T3), and template-based sentences (T4). These different annotation forms are designed to support diverse research objectives in vision-language learning, enabling both expressive semantic understanding and controllable representation analysis.

\begin{figure*}[t]
	\centering
	\includegraphics[width=\textwidth]{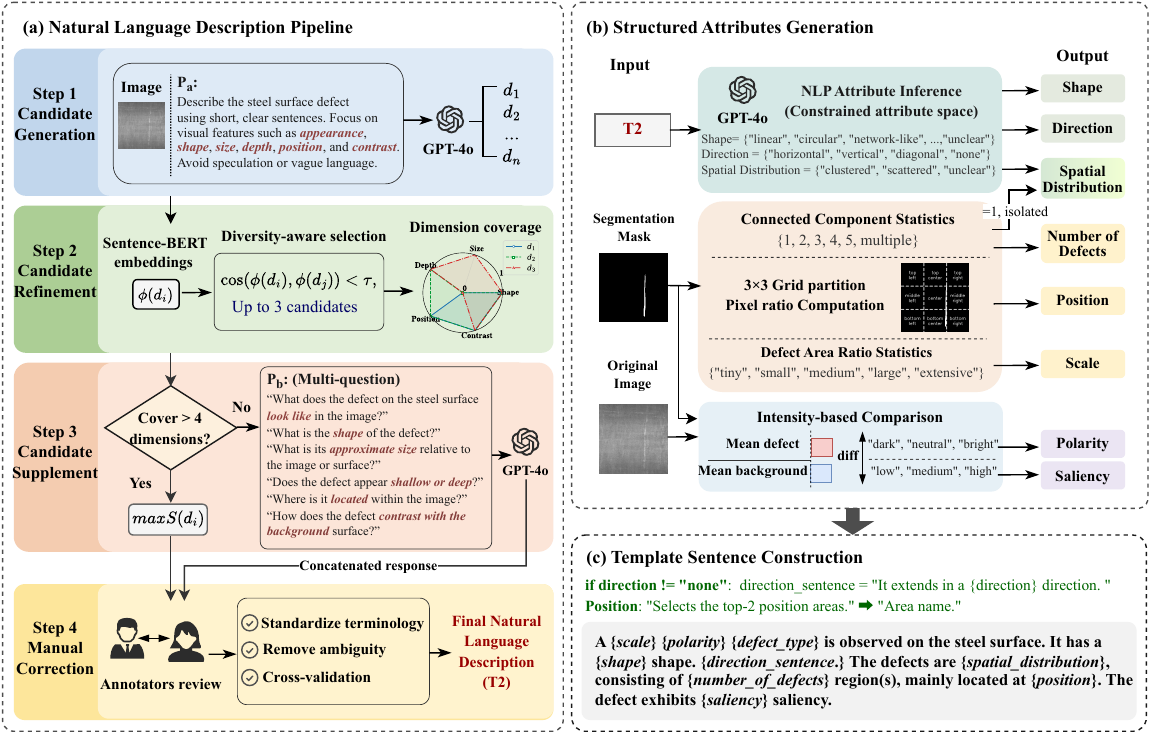}  
	\caption{Sample-level annotation pipeline. 
		The framework produces three text forms for each defect sample: natural language descriptions (T2), structured attributes (T3), and template-based sentences (T4).
		(a) ``T2'' is generated through candidate generation, refinement, supplementation, and manual correction. 
		(b) ``T3'' combines rule-based measurements from images and masks with constrained semantic inference to produce a fixed nine-field attribute representation. 
		(c) ``T4'' linearizes structured attributes into a standardized sentence for controllable vision-language learning.}
	\label{fig:sample_level_annotation}
\end{figure*} 

\subsubsection{Natural Language Description}
We design an automated, structured pipeline that generates high-quality natural-language descriptions. As shown in Fig.~\ref{fig:sample_level_annotation}(a), it comprises candidate generation, refinement, supplementation, and manual correction.  The process is guided by a predefined semantic space covering \textit{appearance}, \textit{shape}, \textit{size}, \textit{depth}, \textit{position}, and \textit{contrast}, which capture key visual perceptual factors in industrial defect inspection and ensure both coverage and consistency.

\textbf{Step 1: Candidate Generation}. For each image $I$, an open-ended instruction prompt $P_a$ guides GPT-4o~\cite{hurst2024gpt} to produce multiple natural-language descriptions of visible defect characteristics. We adopt a relatively high sampling temperature to encourage linguistic diversity and reduce template bias. Multiple sampling rounds generate $n$ candidate descriptions $\{d_1, d_2, \dots, d_n\}$ ($n=4$, \textit{temperature}=0.9, \textit{top\_p}=0.9, \textit{max\_tokens}=80). This stage prioritizes semantic diversity so that different visual aspects of the defect may be expressed across candidate descriptions.

\textbf{Step 2: Candidate Refinement}. 
To remove redundancy while preserving diversity, we perform a similarity-based selection using Sentence-BERT~\cite{reimers2019sentence} embeddings $\phi(d_i)$. Starting from an initial candidate, each subsequent description is retained only if its maximum cosine similarity to the selected set is below $\tau=0.9$, yielding a subset $\mathcal{D}'$ with at most three candidates (i.e., $|\mathcal{D}'|\leq 3$) and preventing description collapse.

To assess whether descriptions capture meaningful defect characteristics, we evaluate semantic coverage over five explicit dimensions. Since \textit{appearance} is included in each description as a global summary and is not suitable for stable keyword-based evaluation, we focus on five well-defined dimensions: \textit{shape}, \textit{size}, \textit{depth}, \textit{position}, and \textit{contrast}. Each retained description $d_i \in \mathcal{D}'$ is encoded as a 5-bit binary vector $\mathbf{b}_i = [b_{1}, b_{2}, b_{3}, b_{4}, b_{5}] \in \{0,1\}^5$, where $b_k=1$ indicates that the $k$-th dimension is mentioned, and $0$ otherwise. Each dimension is associated with a predefined keyword set, such as ``\textit{shape}'' (circle, irregular, linear, etc.). This formulation enables explicit quantification of semantic completeness and promotes consistent annotation structure. 

To balance coverage and diversity, each candidate is assigned a score. The semantic dissimilarity of a description $d_i$ is defined as:
\begin{equation}
	D(d_i) = 1 - \frac{1}{|\mathcal{D}'|-1} \sum_{j \ne i} \cos\big(\phi(d_i), \phi(d_j)\big),
	\label{eq:d_similarity}
\end{equation}
\noindent
and the final score is:
\begin{equation}
	S(d_i) =
	\begin{cases}
		\lambda_1 \cdot \dfrac{\|\mathbf{b}_i\|_1}{5}
		+ \lambda_2 \cdot D(d_i), & |\mathcal{D}'| > 1, \\[6pt]
		\lambda_1 \cdot \dfrac{\|\mathbf{b}_i\|_1}{5}, & |\mathcal{D}'| = 1,
	\end{cases}
	\label{eq:score}
\end{equation}
\noindent
with $\lambda_1=0.6$ and $\lambda_2=0.4$. The highest-scoring candidate is selected as the final annotation, ensuring both semantic coverage and diversity.

\textbf{Step 3: Candidate Supplement}. If no candidate covers at least four of the five dimensions, a structured multi-question prompt $P_b = \{q_1, \dots, q_6\}$ is used to explicitly target each visual aspect. Unlike $P_a$, which emphasizes diversity and naturalness, $P_b$ ensures completeness. The responses are concatenated to form a comprehensive description that balances diversity with dimensional coverage.

\textbf{Step 4: Manual Correction}. To ensure the reliability of the generated descriptions, two annotators conducted approximately 275 hours of cross-validation to standardize industrial terminology and remove ambiguity. Approximately 65\% of the descriptions were manually revised or corrected. The further quality analysis in Appendix~\ref{subsec:T2Annotation} shows that the ``T2'' annotations are both semantically complete and linguistically well-structured.

\subsubsection{Structured Attributes} 
To provide more explicit and controllable semantic representations, we construct a structured attribute form (T3) for each defect sample. Unlike ``T2'', which mainly describes overall visual appearance in natural language, ``T3'' encodes more fine-grained defect characteristics using predefined attribute fields and measurable visual statistics. Each sample is represented as a nine-field tuple:
\textit{``Defect type: \{\}. Shape: \{\}. Direction: \{\}. Spatial Distribution: \{\}. Number of Defects: \{\}. Position: \{\}. Scale: \{\}. Polarity: \{\}. Saliency: \{\}''}.

The attributes are derived through a hybrid pipeline combining deterministic computation and constrained semantic inference. For attributes that can be directly estimated from segmentation masks and image statistics, we apply rule-based computation. As shown in Fig.~\ref{fig:sample_level_annotation}(b), \textit{number of defects} is obtained via connected-component analysis; \textit{position} is represented using a $3\times3$ spatial grid with defect-pixel ratios; \textit{scale} is computed from the defect-area ratio and discretized into five levels from \textit{tiny} to \textit{extensive}; \textit{polarity} is defined by the intensity difference between defect and background regions; and \textit{saliency} is derived from the magnitude of this intensity contrast. \textit{Spatial distribution} is determined using a combination of connected-component statistics and constrained LLM prediction. For attributes requiring semantic interpretation, such as \textit{shape} and \textit{direction}, we employ an LLM to perform constrained classification over predefined candidate sets, using ``T2'' as semantic context. 

Additional details on the structured attribute definitions and discretization rules are provided in Appendix~\ref{subsec:T3_a}. This hybrid design ensures that most attributes are derived from deterministic and reproducible computations, while semantic attributes remain consistent and reduce ambiguity through constrained prediction. Such a design is critical for reliable evaluation in vision-language learning.

\subsubsection{Template Sentence}
Based on the structured attributes, we further construct a template-based sentence representation (T4) to produce consistent text. Specifically, the attributes are linearized into the following sentence pattern:
\textit{A \{scale\} \{polarity\} \{defect\_type\} is observed on the steel surface. It has a \{shape\} shape. \{direction\_sentence\}. The defects are \{spatial\_distribution\}, consisting of \{number\_of\_defects\} region(s), mainly located at \{position\}. The defect exhibits \{saliency\} saliency.}

To ensure linguistic conciseness while preserving spatial informativeness, candidate positions are ranked by area ratios derived from segmentation masks, and the top-2 positions are selected and combined into a coherent phrase (e.g., ``center and left-center''). The directional clause is included only when a clear orientation is present, avoiding ambiguous or unreliable descriptions. Compared with ``T2'', ``T4'' preserves the core semantic attributes while reducing linguistic variability, resulting in a more standardized and controllable text representation for downstream analysis.

\subsection{Pixel-Level Annotation}
We provide pixel-level binary masks for all defect images, where annotated foreground regions correspond to visible defect areas.  Some masks are directly adopted from publicly available datasets. Others are generated using a semi-automatic process, ISAT ~\cite{ISAT_with_segment_anything}, which combines segmentation models with manual refinement to ensure efficiency and annotation quality.

\section{Benchmark and Experiments}
\label{sec:exper}

\subsection{Experimental Setup}
We build a multimodal benchmark on SteelDefectX to evaluate how different forms (``T0''-``T4'') of textual supervision affect vision-language learning for steel surface defect analysis. It covers three primary tasks: vision-language classification, vision-language segmentation, and cross-dataset transfer. To further examine the semantic utility of our annotations, we also conduct structured attribute ablation, vision-language retrieval, text-guided localization, and image-text similarity analysis. All experiments follow the official train/validation split of SteelDefectX with a ratio of 7:3. Additional details on experimental compute resources are provided in Appendix~\ref{subsec:Exper_compute_resources}.

\textbf{Evaluation Metrics.} For classification, we report overall accuracy (Acc) and mean class accuracy (mAcc). For segmentation and text-guided localization, we use AUROC~\cite{defard2021padim}, F1-max~\cite{wang2025normalabnormal}, and IoU~\cite{ni2026towards}, computed from pixel-wise response maps. For vision-language retrieval~\cite{Wang2024SkyScript, jiang2025cross}, we report Recall@1, Recall@5, and Recall@10. We also report the average image-text similarity (ImgTxtSim) score to measure global semantic alignment between visual and textual representations.

\subsection{Vision-Language Classification}
\label{Subsec: vl_classification}
This task follows a vision-language matching paradigm based on CLIP ~\cite{radford2021learning}, which employs a vision encoder to extract visual features and a text encoder to encode textual representations of all possible label templates. 
The model then computes the similarity between visual and textual features, and selects the label whose textual feature has the highest similarity to the visual feature as the prediction. We evaluate five representative CLIP variants, including CLIP~\cite{radford2021learning}, OpenCLIP~\cite{Cherti2023Reproducible}, EVA-CLIP~\cite{sun2023eva}, Long-CLIP~\cite{Zhang2024LongCLIP}, and FG-CLIP~\cite{xie2025fgclip}. All models are trained following CLIP-Adapter~\cite{gao2024clip}, which adapts a pretrained CLIP model for downstream image classification by introducing lightweight residual adapters into the visual encoders. We train for 20 epochs using the Adam optimizer with a learning rate of 1e-4 and bidirectional cross-entropy loss. The batch size is set to 16 for training and 32 for validation. In each experiment, the training and evaluation settings are denoted by ``Tx-Ty'', where ``Tx'' corresponds to the training text and ``Ty'' to the evaluation text.

\begin{table*}[t]
	\centering
	\small
	\caption{Vision-language classification results under different training text representations, with evaluation performed using ``T0'' and ``T1''. Best results are in bold.}
	\label{tab:vl_classification_main}
	\resizebox{\linewidth}{!}{
	\begin{tabular}{ll|cc|cc|cc|cc|cc|cc}
		\toprule
		\multirow{2}{*}{Model} & \multirow{2}{*}{Backbone} 
		& \multicolumn{2}{c|}{T2-T0} 
		& \multicolumn{2}{c|}{T2-T1} 
		& \multicolumn{2}{c|}{T3-T0} 
		& \multicolumn{2}{c|}{T3-T1} 
		& \multicolumn{2}{c|}{T4-T0} 
		& \multicolumn{2}{c}{T4-T1} \\
		\cmidrule(lr){3-4} \cmidrule(lr){5-6} \cmidrule(lr){7-8} \cmidrule(lr){9-10} \cmidrule(lr){11-12} \cmidrule(lr){13-14}
		&
		& Acc & mAcc 
		& Acc & mAcc 
		& Acc & mAcc 
		& Acc & mAcc 
		& Acc & mAcc 
		& Acc & mAcc \\
		\midrule
		CLIP~\cite{radford2021learning} & ViT-B/16 & 82.96 & 82.42 & 76.94 & 73.13 & 90.75 & 88.08 & 77.62 & 79.21 & 85.80 & 85.69 & 74.96 & 75.66 \\
		OpenCLIP~\cite{Cherti2023Reproducible} & ViT-B/16 & 87.13 & 83.99 & 78.06 & 75.24 & 85.20 & 83.40 & 74.70 & 68.07 & 86.19 & 86.01 & 78.23 & 77.03 \\
		EVA-CLIP~\cite{sun2023eva} & ViT-B/16 & 85.54 & 80.53 & 78.79 & 73.48 & 83.69 & 85.79 & 78.31 & 78.84 & 86.19 & 85.42 & 73.49 & 70.67 \\
		Long-CLIP~\cite{Zhang2024LongCLIP} & ViT-B/16 & 88.25 & 85.41 & 82.83 & 79.11 & 90.23 & 91.27 & 82.92 & 81.96 & 89.50 & 89.35 & 78.23 & 77.32 \\
		FG-CLIP~\cite{xie2025fgclip} & ViT-B/16 & 84.77 & 81.66 & 80.29 & 77.24 & 89.37 & 88.07 & 84.55 & 84.60 & 87.31 & 86.50 & 77.67 & 77.92 \\
		\midrule
		CLIP~\cite{radford2021learning} & ViT-L/14 & 82.66 & 82.24 & 74.53 & 71.51 & 90.71 & 88.94 & 78.10 & 77.91 & 84.04 & 79.65 & 67.64 & 63.59 \\
		OpenCLIP~\cite{Cherti2023Reproducible} & ViT-L/14 & 88.55 & 87.27 & 76.76 & 77.29 & 86.40 & 83.30 & 76.12 & 78.77 & 90.10 & 88.70 & 78.87 & 79.66 \\
		EVA-CLIP~\cite{sun2023eva} & ViT-L/14 & 87.48 & 84.32 & 77.41 & 74.62 & 89.80 & 87.26 & 83.35 & 82.24 & 88.34 & 85.99 & 79.13 & 77.54 \\
		Long-CLIP~\cite{Zhang2024LongCLIP} & ViT-L/14 & \textbf{93.63} & \textbf{92.56} & \textbf{92.30} & \textbf{86.79} & \textbf{94.45} & \textbf{93.66} & \textbf{93.12} & \textbf{90.70} & \textbf{93.03} & \textbf{91.47} & \textbf{89.50} & \textbf{84.60} \\
		FG-CLIP~\cite{xie2025fgclip} & ViT-L/14 & 92.77 & 90.32 & 87.91 & 84.70 & 93.42 & 91.93 & 89.85 & 89.68 & 89.46 & 89.34 & 88.64 & 85.47 \\
		\bottomrule
	\end{tabular}}
\end{table*}

As shown in Table~\ref{tab:vl_classification_main}, Long-CLIP achieves the best overall performance, with ViT-L/14 reaching 94.45\% Acc and 93.66\% mAcc under the ``T3-T0'' setting, while FG-CLIP also performs competitively. Across textual representations, ``T3'' consistently achieves strong performance, suggesting that structured attributes provide more explicit and less ambiguous supervision than free-form descriptions or templates. Furthermore, we observe a noticeable performance drop when evaluating with ``T1'', which is attributed to the semantic gap between the training and inference texts. These results highlight that SteelDefectX enables systematic analysis of how textual structure affects semantic alignment and model robustness.

\subsection{Vision-Language Segmentation}
\label{Subsec: vl_segmentation}
\begin{figure*}[t]
	\centering
	\begin{minipage}{0.52\linewidth}
		\centering
		\captionsetup{type=table}
		\caption{Vision-language segmentation results on SteelDefectX. Fixed text prompts are used for Long-CLIP and WinCLIP, while AnomalyCLIP uses learned prompts. Best results are in bold.}
		\vspace{0pt}
		\label{tab:vl_segmentation_main}
		\resizebox{\linewidth}{!}{
			\begin{tabular}{lc ccc}
				\toprule
				Model  & Text & AUROC & F1-max & IoU \\
				\midrule
				\multirow{5}{*}{\makecell[c]{Long-CLIP~\cite{Zhang2024LongCLIP}\\ViT-L/14}} & T0 & 78.96 & 40.46 & 21.92 \\
				& T1 & 70.10 & 30.67 & 14.14 \\
				& T2 & 77.89 & 39.53 & 21.04 \\
				& T3 & 72.49 & 33.42 & 15.92 \\
				& T4 & 75.33 & 37.08 & 18.63 \\
				\midrule
				\multirow{5}{*}{\makecell[c]{WinCLIP~\cite{Jeong2023WinCLIP} \\ ViT-B/16+}} & T0 & 71.51 & 30.89 & 16.56 \\
				& T1 & 71.87 & 31.16 & 16.68 \\
				& T2 & 71.68 & 31.02 & 16.63 \\
				& T3 & 71.13 & 29.75 & 14.74 \\
				& T4 & 71.66 & 31.05 & 16.51 \\
				\midrule
				\makecell[c]{AnomalyCLIP~\cite{zhou2024anomalyclip} \\ ViT-L/14} & learned & \textbf{88.21} & \textbf{53.36} & \textbf{37.49} \\
				\bottomrule
		\end{tabular}}
	\end{minipage}
	\hfill
	\begin{minipage}{0.46\linewidth}
		\centering
		\includegraphics[width=\linewidth]{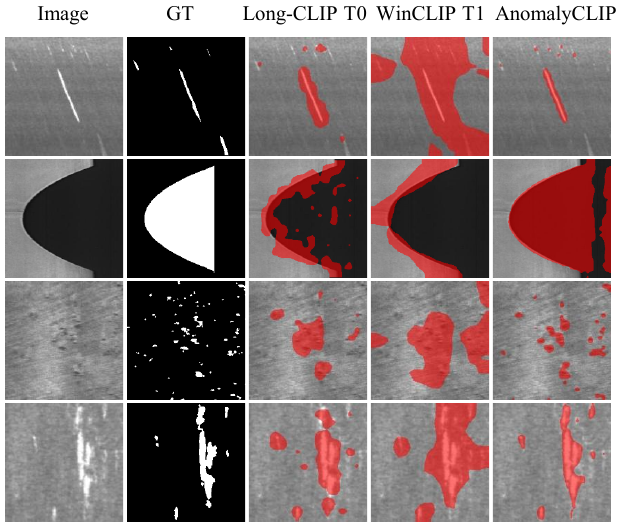}
		\vspace{-8pt}
		\caption{Qualitative comparison of vision-language segmentation results.}
		\label{fig:seg_c}
	\end{minipage}
\end{figure*}

This task evaluates vision-language models at the pixel level by matching dense visual features with text embeddings. We evaluate three models: WinCLIP~\cite{Jeong2023WinCLIP}, AnomalyCLIP~\cite{zhou2024anomalyclip}, and Long-CLIP Adapter~\cite{Zhang2024LongCLIP}. For each text setting, all descriptions within the same defect category are aggregated into a shared text library and encoded into class prototypes. Although ``T2''-``T4'' originate from sample-level annotations, localization is still performed using class-level text guidance. The visual encoder extracts dense image features, and similarity between local visual tokens and class prototypes produces localization maps, which are compared against ground-truth segmentation masks. WinCLIP and AnomalyCLIP follow their original evaluation pipelines, while their text inputs are replaced with the corresponding SteelDefectX text variants whenever fixed-text evaluation is performed.

As shown in Table~\ref{tab:vl_segmentation_main}, AnomalyCLIP with learned prompts achieves the best overall performance (AUROC: 88.21\%, F1-max: 53.36\%, IoU: 37.49\%), confirming the benefit of task-specific prompt optimization. Long-CLIP consistently outperforms WinCLIP under fixed-text settings. Notably, richer textual descriptions (``T2''-``T4'') do not consistently improve segmentation performance under class-level aggregation. In particular, ``T0'' achieves the best results for Long-CLIP, while ``T2'' remains competitive and ``T3''/``T4'' perform slightly worse. This suggests that fine-grained semantic information may be diluted when aggregated into class-level prompts. It further highlights a limitation of current vision–language segmentation models, which are constrained to class-level prompt segmentation. Fig.~\ref{fig:seg_c} provides qualitative comparisons using the best-performing text setting for each model. These results highlight the role of textual structure in spatial grounding and demonstrate the value of SteelDefectX as a benchmark for text-aware localization research. To further analyze sample-level alignment, we provide 1-to-1 text-guided segmentation and image-text similarity experiments in Section \ref{sec:further_analysis}. 

\subsection{Cross-Dataset Transfer}
\label{Subsec: cross_dataset}
We evaluate cross-dataset transfer to examine whether different textual representations learned on SteelDefectX generalize to unseen defect datasets~\cite{liATPrompt2025}. For the experiment, we adopt Long-CLIP~\cite{Zhang2024LongCLIP} as the base model and evaluate it on two external datasets: MSD-Cls~\cite{xiao2022graph} (aluminum surface defects) and CGFSDS-9~\cite{song2024mfanet} (seamless steel tube defects). The original model is directly evaluated as a zero-shot baseline, and then compared with models trained on SteelDefectX using different textual annotations (``T0''-``T4''). To ensure consistent semantics across materials, a unified template (``\textit{A photo of surface defect: [classname]}'') is used during evaluation. Training settings follow Section~\ref{Subsec: vl_classification}.

\begin{table}[t]
	\centering
	\caption{Cross-dataset transfer results (\% Acc) of Long-CLIP Adapter trained on SteelDefectX and evaluated on two unseen defect datasets. Best results are in bold.}
	\label{tab:cross_dataset_transfer}
	\resizebox{0.85\linewidth}{!}{
		\begin{tabular}{l l c c c c c c}
			\toprule
			Dataset & Backbone & Zero-shot & T0 & T1 & T2 & T3 & T4 \\
			\midrule
			Aluminum~\cite{xiao2022graph} & ViT-B/16 & 10.22 & 15.59 & 18.01 & 22.58 & 19.35 & \textbf{23.12} \\
			& ViT-L/14 & 8.60  & 12.90 & 20.43 & \textbf{29.03} & 12.37 & 15.59 \\
			\midrule
			Seamless Steel Tubes~\cite{song2024mfanet} & ViT-B/16 & 31.05 & 31.51 & 33.33 & \textbf{43.38} & 37.90 & 31.05 \\
			& ViT-L/14 & 25.11 & 28.31 & 33.79 & \textbf{40.18} & 28.77 & 27.40 \\
			\bottomrule
	\end{tabular}}
\end{table}

As shown in Table~\ref{tab:cross_dataset_transfer}, training on SteelDefectX consistently improves performance over the zero-shot baseline, indicating effective cross-dataset transfer. Among all text settings, ``T2'' achieves the best or near-best results across most settings, indicating that free-form natural language descriptions generalize better than structured attributes (T3) and template-based text (T4). This suggests that semantic richness and linguistic flexibility play a key role in cross-dataset transfer, while relatively fixed representations may limit generalization. In summary, these results highlight the importance of text design in multimodal transfer and demonstrate that SteelDefectX enables systematic study of generalization across datasets in industrial defect analysis.

\subsection{Analysis and Discussion}
\label{sec:further_analysis}
To better understand how different textual representations affect vision-language alignment, we conduct additional analyses, including structured attribute ablation, vision-language retrieval, and text-guided localization. These analyses aim to reveal how semantic structure and expressiveness influence model behavior beyond standard benchmark tasks.

\begin{table}[t]
	\small
	\centering
	
	\begin{minipage}[t]{0.41\textwidth}
		\centering
		\caption{Ablation of structured attributes (T3) using Long-CLIP Adapter (ViT-L/14). The model is trained with the defect type and one attribute, and evaluated Acc(\%) under ``T0'' and ``T1''.} 
		\vspace{3pt}
		\label{tab:structured_attribute_ablation}
		\resizebox{\textwidth}{!}{
		\begin{tabular}{lcc}
			\toprule
			Attribute & T0 & T1 \\
			\midrule
			Defect type & 90.23 & 87.18 \\
			\midrule
			+Shape & 91.70 & 85.76 \\
			+Direction & 92.99 & 86.75 \\
			+Spatial Distribution & \textbf{95.52} & 90.02 \\
			+Number of Defects & 94.45 & 87.87 \\
			+Position & 92.60 & 87.05 \\
			+Scale & 93.12 & 80.03 \\
			+Polarity & 95.14 & 87.52 \\
			+Saliency & 94.66 & 88.94 \\
			\midrule
			T3 & 94.45 & \textbf{93.12} \\
			T4 & 93.03 & 89.50 \\
			\bottomrule
		\end{tabular}}
	\end{minipage}
	\hfill
	\begin{minipage}[t]{0.56\textwidth}
		
		\begin{minipage}[t]{0.98\textwidth}
			\centering
			\caption{Vision-language retrieval results of Long-CLIP Adapter (ViT-L/14) under different texts. Best results are in bold.}
			\label{tab:vl_retrieval_longclip}
			\resizebox{0.98\textwidth}{!}{
			\begin{tabular}{lcccccc}
				\toprule
				\multirow{2}{*}{Text} & \multicolumn{3}{c}{Image to Text} & \multicolumn{3}{c}{Text to Image} \\
				\cmidrule(lr){2-4} \cmidrule(lr){5-7}
				& R@1 & R@5 & R@10 & R@1 & R@5 & R@10 \\
				\midrule
				T2 & 2.80 & 11.19 & 20.01 & 3.23 & 11.02 & 18.98 \\
				T3 & 5.77 & \textbf{21.86} & 35.93 & 4.91 & 19.54 & 32.36 \\
				T4 & \textbf{5.85} & 21.30 & \textbf{36.06} & \textbf{5.64} & \textbf{20.44} & \textbf{33.99} \\
				\bottomrule
			\end{tabular}}
		\end{minipage}
		
		\vspace{0.5em}
		
		\begin{minipage}[t]{0.98\textwidth}
			\centering
			\caption{Text-guided segmentation and image-text similarity results of FG-CLIP (ViT-L/14) under different texts. Best results are in bold.}
			\label{tab:fgclip_text_guided_localization}
			\resizebox{0.85\textwidth}{!}{
			\begin{tabular}{ccccc}
				\toprule
				Text & AUROC & F1-max & IoU & ImgTxtSim \\
				\midrule
				T0 & 44.80 & 18.04 & 9.92 & 0.1144 \\
				T1 & 53.66 & 18.19 & 9.74 & 0.1261 \\
				T2 & \textbf{58.54} & \textbf{20.48} & 10.03 & 0.1843 \\
				T3 & 58.07 & 20.06 & \textbf{10.14} & 0.1897 \\
				T4 & 54.53 & 18.43 & 9.45 & \textbf{0.1997} \\
				\bottomrule
			\end{tabular}}
		\end{minipage}
		
	\end{minipage}
	
\end{table}

\textbf{Structured Attributes.} We analyze the contribution of each structured attribute (T3) to vision-language classification using Long-CLIP Adapter (ViT-L/14)~\cite{Zhang2024LongCLIP}. The model is trained with prompts that combine a defect type with a single attribute and is evaluated under ``T0'' and ``T1''. As shown in Table~\ref{tab:structured_attribute_ablation}, \textit{Spatial Distribution} performs best among individual attributes (95.52 / 90.02), followed by \textit{Polarity} and \textit{Saliency}, indicating the importance of spatial layout and contrast cues. Other attributes yield moderate improvements over the defect-type baseline. Performance under ``T1'' is generally lower, suggesting that its richer semantics introduce greater intra-class variability beyond what a single attribute can capture. In contrast, ``T3'' achieves the best performance under ``T1'' (93.12), demonstrating that combining attributes leads to more robust alignment. An attribute-level analysis for dense visual grounding is provided in Appendix~\ref{subsec:sup_attr}.

\textbf{Vision-Language Retrieval.} To evaluate cross-modal alignment under different text forms (``T2''-``T4''), we use the Long-CLIP Adapter (ViT-L/14)~\cite{Zhang2024LongCLIP} for image-text retrieval. Table~\ref{tab:vl_retrieval_longclip} shows that structured representations (T3, T4) consistently outperform free-form descriptions (T2), with ``T4'' achieving the best overall performance. These results indicate that structured representations provide more consistent semantic signals, thereby facilitating more stable cross-modal alignment. In contrast, the higher variability in free-form descriptions introduces greater semantic ambiguity, making image-text matching more difficult.

\textbf{Text-Guided Segmentation.}
To evaluate the effect of different textual representations on fine-grained visual grounding, we conduct text-guided segmentation using FG-CLIP (ViT-L/14)~\cite{xie2025fgclip}. Given an image-text pair, patch-text similarities are computed to produce a spatial response map, which is normalized and used for localization. We compare five text types. As shown in Table~\ref{tab:fgclip_text_guided_localization}, richer texts generally outperform class-name prompts. ``T2'' achieves the best AUROC and F1-max, while ``T3'' yields the highest IoU. Despite achieving the highest image-text similarity, ``T4'' underperforms in localization, indicating that strong global alignment does not necessarily translate into accurate spatial grounding. These results highlight the importance of fine-grained textual descriptions and suggest that overly structured representations may be less effective for spatial grounding.

\section{Conclusion}
We introduce SteelDefectX, the first vision-language dataset with multi-form textual annotations for steel surface defect analysis. The dataset provides diverse textual representations, along with unified benchmarks for classification, segmentation, and cross-dataset transfer, enabling a systematic study of semantic alignment and generalization in industrial vision-language learning. Experimental results highlight the impact of text representation design on industrial vision-language models. In particular, structured annotations facilitate stable alignment, while free-form natural language descriptions improve transferability and fine-grained defect grounding. We expect SteelDefectX to serve as a valuable resource for advancing industrial multimodal learning and developing more interpretable and generalizable inspection systems. The dataset may also facilitate future research in zero- and few-shot recognition, text-guided generation, and industrial multimodal reasoning.

%% file: Appendix.tex
\section{Dataset Construction Details}
\label{sec:sup_a}
The source datasets (NEU~\cite{song2013noise}, GC10~\cite{lv2020deep}, X-SDD~\cite{feng2021x}, and S3D~\cite{cui2024efficient}) differ in annotation granularity, image quality, and naming conventions. Some categories share similar semantic meanings but exhibit distinct visual patterns, whereas others appear visually similar yet differ in their definitions. To address these inconsistencies, we systematically review and reconcile all labels through visual inspection and expert verification. Specifically, ``Inclusion'' in NEU and GC10 refers to comparable defects and is merged, whereas ``Inclusion'' in X-SDD represents a visually distinct slag-related defect and is redefined as ``Slag inclusion.'' Similarly, ``Scratches'' in NEU and ``Bright scratch'' in S3D are consolidated as ``Bright scratch,'' while ``Scratches'' in X-SDD and ``Pressure scratch'' in S3D are unified as ``Dark scratches.'' Through this human-guided consolidation, semantically and visually related subclasses are integrated into a standardized dataset comprising 25 categories. This consolidation process ensures semantic clarity and visual diversity, providing a unified and interpretable foundation for subsequent class-level and sample-level description generation.

\begin{figure*}[h]
	\centering
	\includegraphics[width=\textwidth]{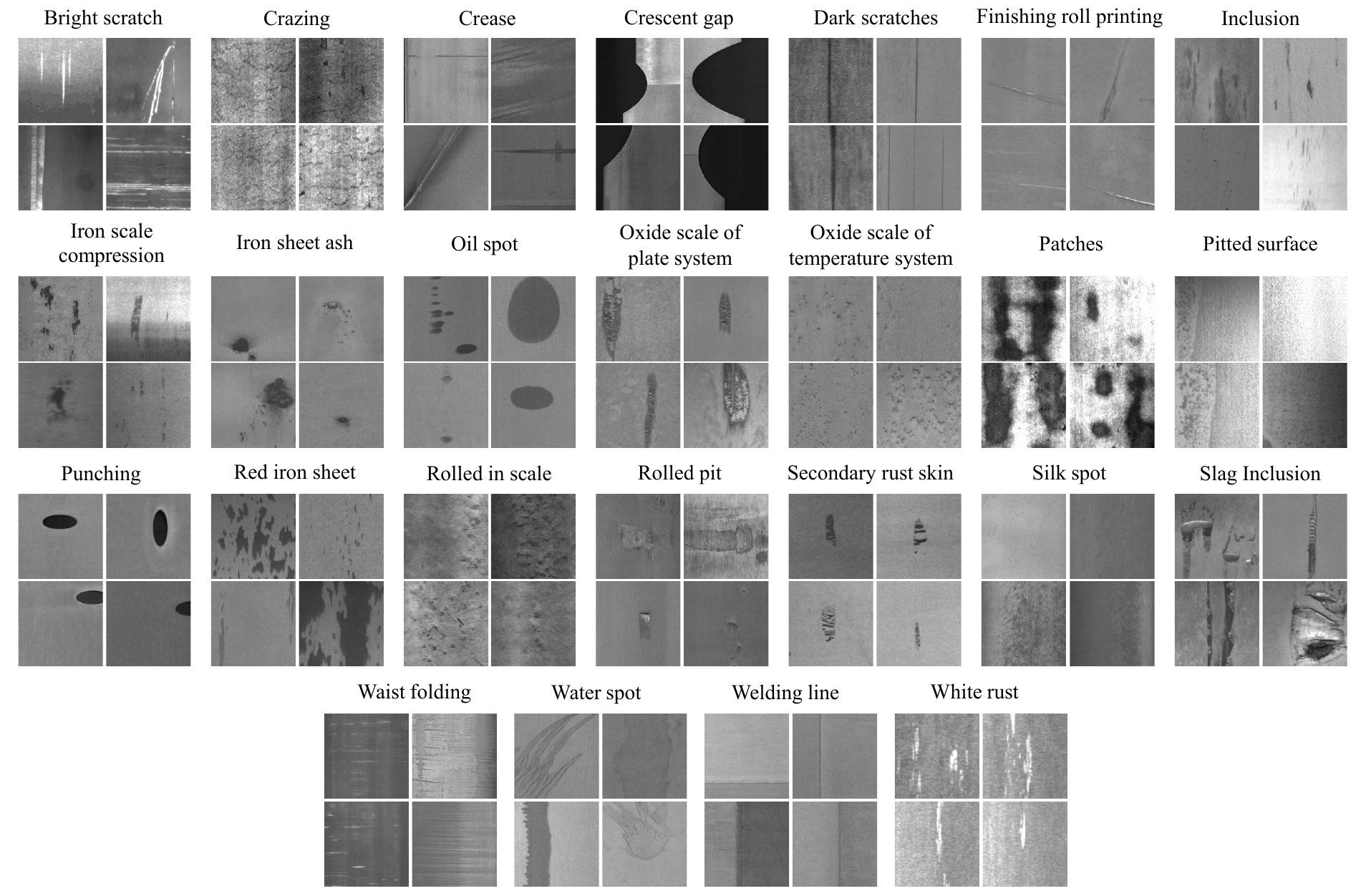}
	\caption{Representative samples of each defect category in the SteelDefectX. For each class, four images are shown to illustrate intra-class variations in appearance, texture, and morphology. The figure highlights substantial diversity in defect scale, shape, illumination, and surface conditions, illustrating the visual complexity of real-world steel defect patterns.}
	\label{fig:example}
\end{figure*}

Representative samples of each defect category are shown in Fig. \ref{fig:example} to illustrate the characteristics of the SteelDefectX. For each class, four samples are selected to reflect the intra-class variability in visual appearance and texture. Some categories (e.g., Punching and Rolled in scale) display strong visual consistency due to well-defined structural patterns, whereas others (e.g., Inclusion and Iron scale compression) exhibit substantial intra-class variability due to their differing defect-formation mechanisms. These variations are further influenced by imaging conditions such as illumination, viewing angle, and surface reflectivity. This visualization demonstrates that the dataset encompasses both visually consistent and diverse defect types, establishing a comprehensive foundation for evaluating model robustness and generalization in real industrial scenarios.

\section{Dataset Analysis}
\label{sec: Dataset_Analysis}

\subsection{Class Distribution and Visual Distribution}
\label{subsec:class_visual}
\begin{figure*}[!t]
	\centering
	\includegraphics[width=0.75\textwidth]{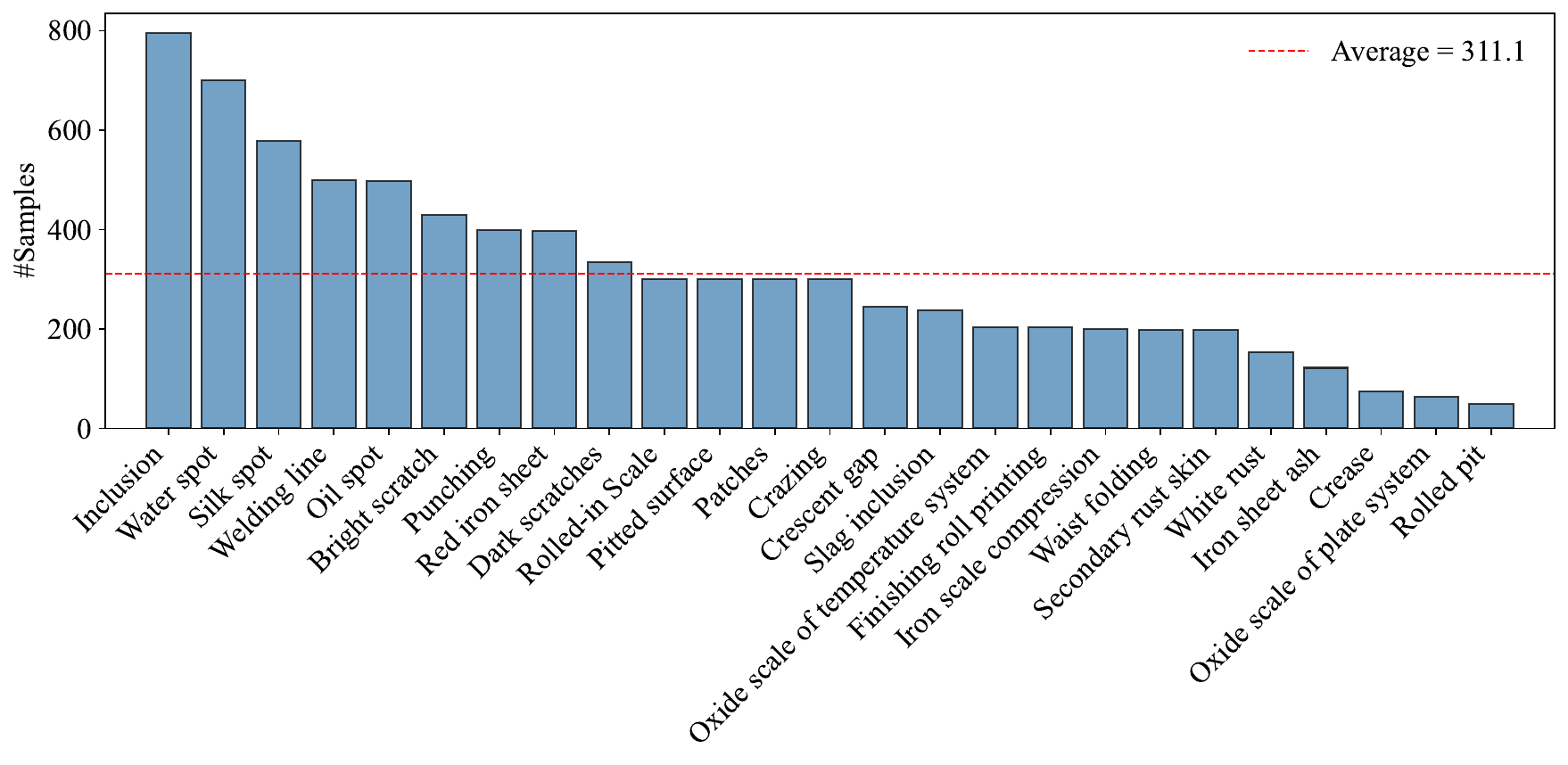}
	\caption{Class distribution of SteelDefectX. The dataset exhibits an imbalanced distribution across 25 defect categories, with sample counts following a log-normal trend. The average number of samples in the dataset is 311. Common defects such as \textit{inclusion} and \textit{water spot} dominate the dataset, whereas rare defects (e.g., \textit{crease} and \textit{rolled pit}) are underrepresented, reflecting real-world variability in steel surface inspection scenarios.}
	\label{fig:fig3}
\end{figure*}

\begin{figure}[t]
	\centering
	\includegraphics[width=0.9\linewidth]{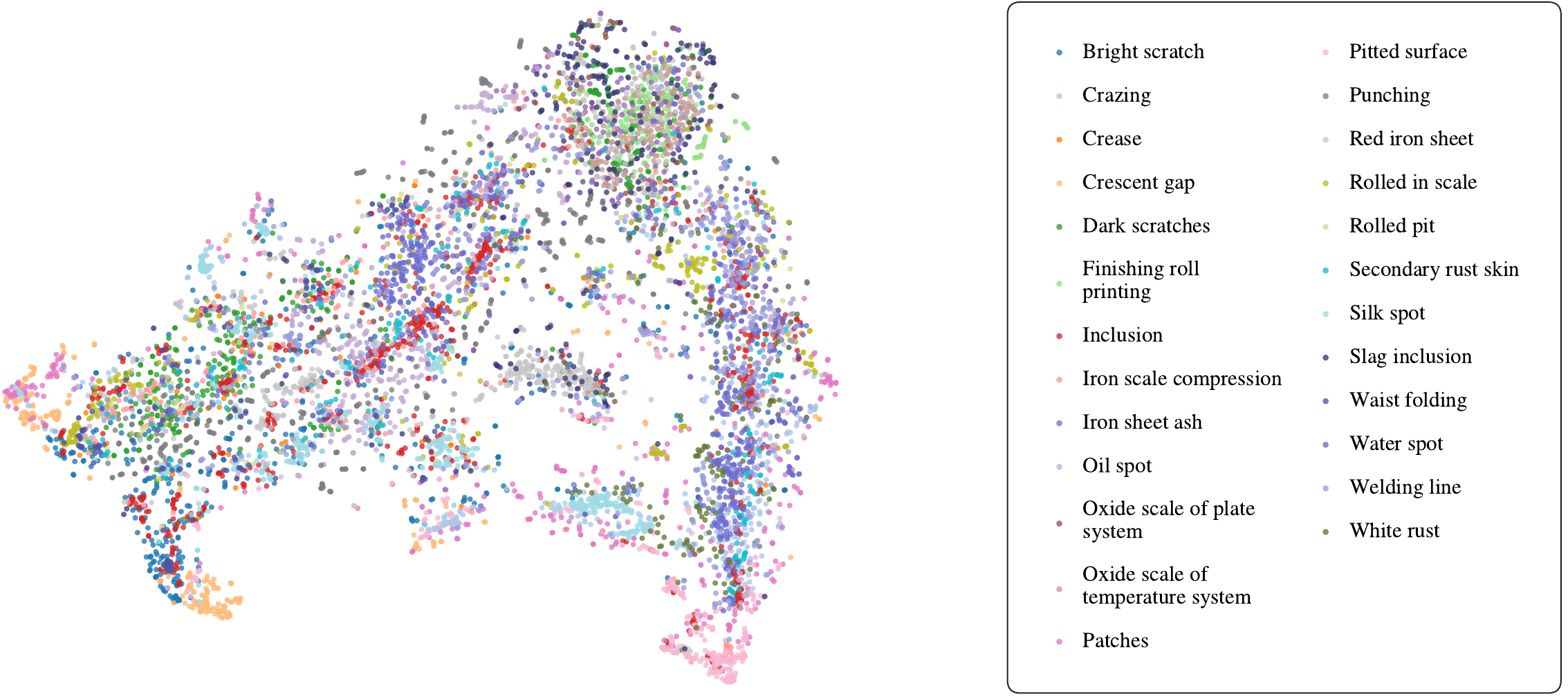}
	\caption{t-SNE visualization of image-level features in SteelDefectX. The distribution reveals substantial intra-class variation and partial inter-class overlap across defect categories, highlighting the visual complexity of industrial defect data.}
	\label{fig:fig4}
\end{figure}

SteelDefectX comprises 7,778 images spanning 25 steel surface defect categories. Each category contains between 50 and 795 samples, reflecting the natural class imbalance commonly observed in manufacturing environments. Fig.~\ref{fig:fig3} visualizes the class distribution, exhibiting a long-tailed pattern consistent with real-world defect occurrence frequencies. This imbalance provides a practical testbed for evaluating few-shot and long-tail learning. To assess visual diversity, Fig.~\ref{fig:fig4} presents t-SNE embeddings of image-level features, revealing substantial intra-class variation and partial inter-class overlap. These characteristics highlight the complexity of defect appearance, which varies across steel types, production processes, and environmental conditions. Such visual heterogeneity poses challenges for reliable defect recognition and underscores the need for robust representation learning.

\subsection{``T2'' Annotation Quality Analysis}
\label{subsec:T2Annotation}

\textbf{Dimension coverage of ``T2'' descriptions.}
To quantitatively evaluate the semantic completeness of ``T2'' descriptions, we consider six predefined dimensions: \textit{appearance}, \textit{shape}, \textit{size}, \textit{depth}, \textit{position}, and \textit{contrast}. Since \textit{appearance} is consistently included in all descriptions, the analysis focuses on the remaining five dimensions. A keyword-based matching strategy is adopted. Representative keywords are curated through corpus-level inspection and iterative refinement. Each description is then examined to determine whether corresponding semantic cues are present. This approach provides a conservative estimate, and the actual coverage is higher.

As shown in Table~\ref{tab:dimension_coverage}, the average coverage reaches 4.39 out of 5, and 94.5\% of the descriptions cover at least four dimensions. This indicates that most descriptions jointly capture key visual attributes. At the dimension level, \textit{shape}, \textit{contrast}, and \textit{position} achieve near-complete coverage (100.0\%, 99.3\%, and 95.4\%), while \textit{size} also maintains high coverage (92.9\%). In contrast, \textit{depth} shows a lower coverage (51.3\%). This is because certain features are inherently difficult to express precisely in natural language. A small degree of trade-off is made during annotation to ensure clarity and reliability. Overall, these results demonstrate that ``T2'' descriptions achieve strong semantic completeness while preserving natural language expressiveness, thereby providing high-quality supervision for vision-language modeling.

\begin{table}[t]
	\centering
	\small
	\caption{Dimension coverage of ``T2''descriptions over five semantic dimensions. ``Coverage $\geq$4'' denotes the proportion of descriptions that cover at least four dimensions.}
	\label{tab:dimension_coverage}
	\begin{tabular}{lccccccc}
		\toprule
		\textbf{Metric} & \textbf{Avg. Coverage} & \textbf{Coverage $\geq$4} & \textbf{Shape} & \textbf{Size} & \textbf{Depth} & \textbf{Position} & \textbf{Contrast} \\
		\midrule
		\textbf{Value} & 4.39 / 5 & 94.5\% & 100.0\% & 92.9\% & 51.3\% & 95.4\% & 99.3\% \\
		\bottomrule
	\end{tabular}
\end{table}

\begin{figure}[t]
	\centering
	\includegraphics[width=\linewidth]{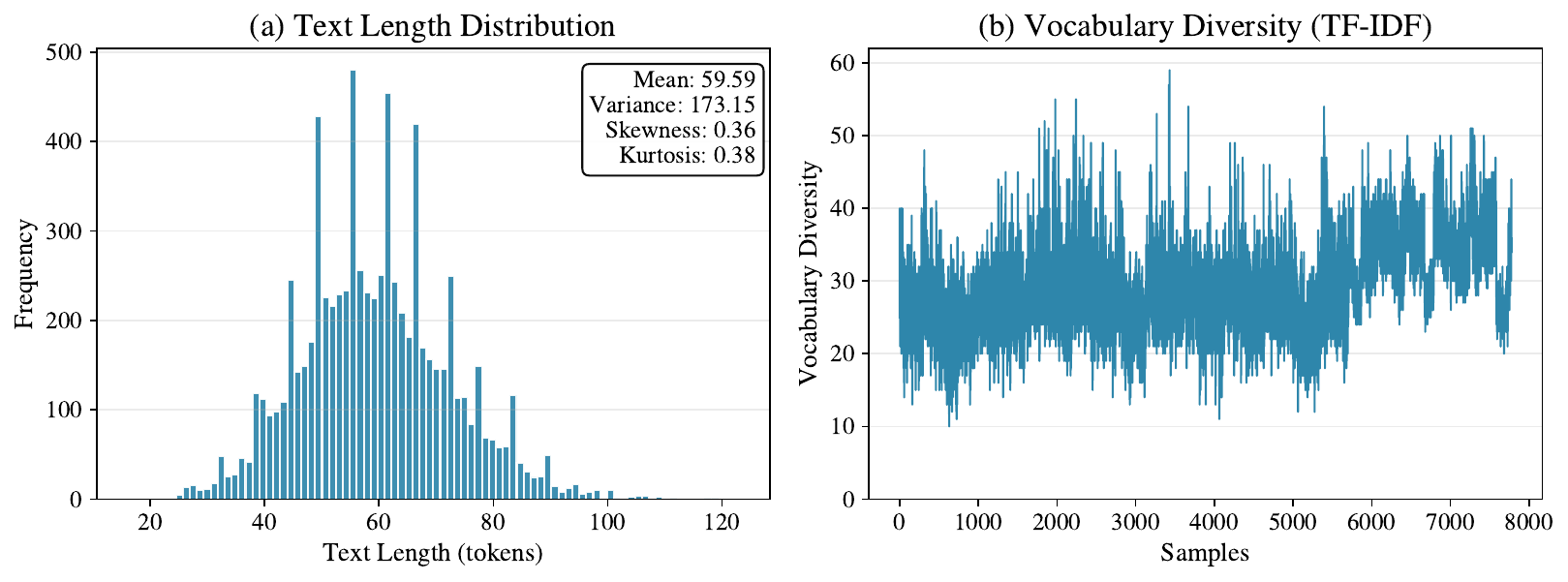}
	\caption{Text statistics of ``T2'' in SteelDefectX. \textbf{(a)} Distribution of description lengths in tokens. The annotations are concentrated around a mean of 59.59 tokens with mild positive skewness, indicating a consistent balance between detail and conciseness. \textbf{(b)} Sample-wise vocabulary diversity measured as the number of non-zero TF-IDF terms after stop-word removal. The relatively stable fluctuation across samples reflects controlled but meaningful lexical variation in the textual annotations.}
	\label{fig:text}
\end{figure}

\textbf{Text Statistics and Vocabulary Diversity Analysis.} To further assess the quality of the ``T2'', we analyze both text length and vocabulary diversity. As shown in Fig.~\ref{fig:text}(a), the text length distribution follows a concentrated and approximately unimodal pattern, with a mean of 59.59 tokens and moderate variance. The distribution exhibits mild skewness and near-normal kurtosis. These statistics indicate that most descriptions are moderately detailed and remain within a relatively stable length range, avoiding both overly short annotations and unnecessarily verbose expressions. 

Fig.~\ref{fig:text}(b) further presents the vocabulary diversity, measured as the number of non-zero features in the TF-IDF representation per sample, corresponding to the number of unique non-stop words in each description. The diversity values exhibit relatively stable variation across samples, suggesting that the descriptions maintain a consistent linguistic structure while preserving meaningful lexical variation. This pattern is consistent with the goal of the proposed generation-and-refinement pipeline, which encourages expressive variation without introducing noise or semantic drift. Therefore, the textual annotations achieve a balance between consistency and expressiveness, which supports robust cross-modal alignment and improved generalization in downstream vision-language tasks.

\subsection{``T3'' Structured Attribute Definition and Discretization}
\label{subsec:T3_a}
The annotations ``T3'' provide a structured and controllable representation of the characteristics of the defects by integrating rule-based measurements with constrained semantic inference. These attributes are designed to balance semantic expressiveness and reproducibility, enabling consistent vision-language alignment across samples.

\begin{figure}[t]
	\centering
	\includegraphics[width=\linewidth]{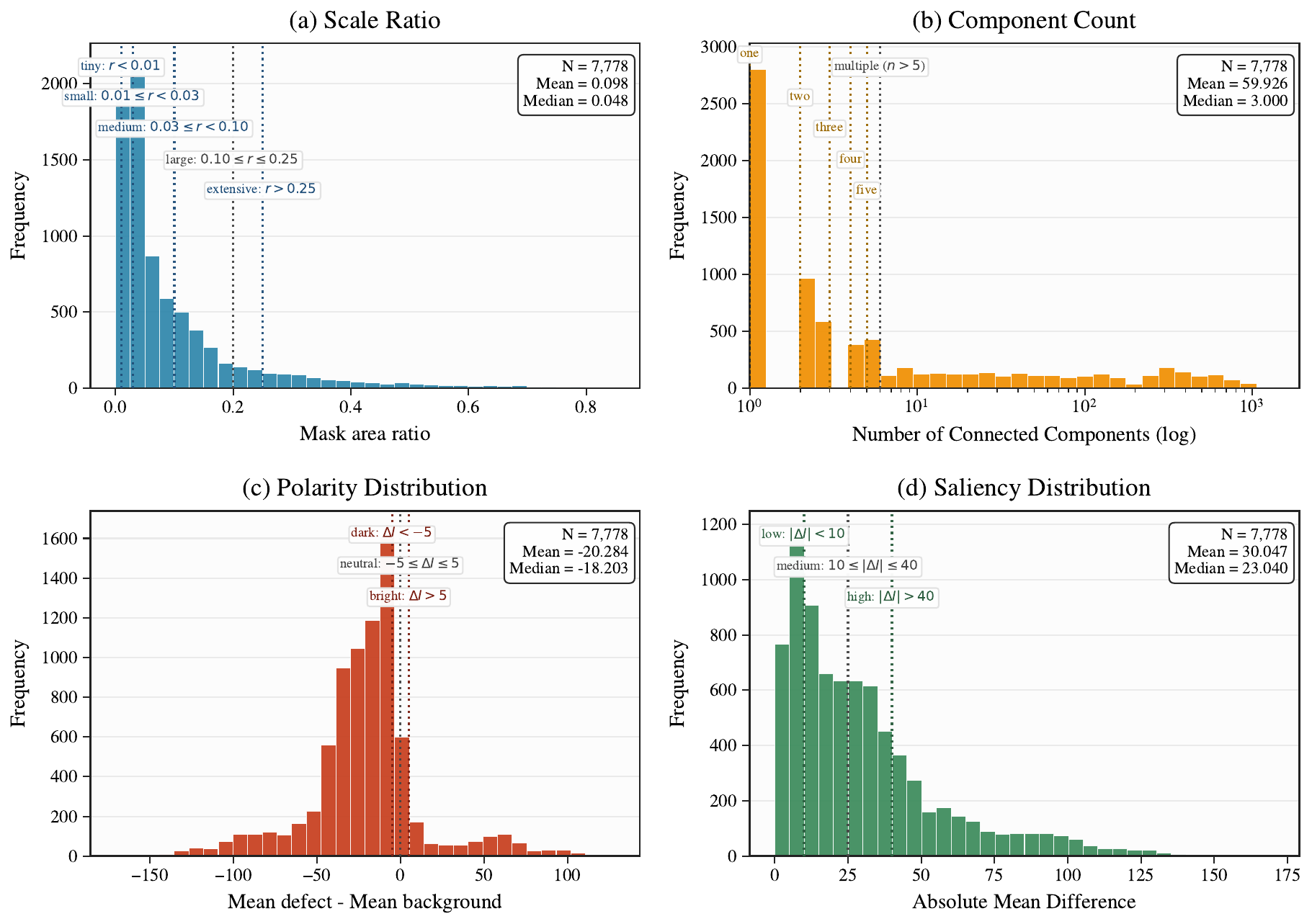}
	\caption{Distribution of rule-based attributes in ``T3'' for the SteelDefectX dataset. Vertical dashed lines indicate discretization thresholds in the structured-attribute framework. \textbf{(a)} Scale ratio, defined as the defect-mask area divided by the full image area. \textbf{(b)} Number of connected components, plotted on a log-scaled horizontal axis. \textbf{(c)} Polarity, measured as the mean intensity difference between the defect region and the background. \textbf{(d)} Saliency, defined as the absolute mean intensity difference.}
	\label{fig:attribute_distribution_overview}
\end{figure}

\begin{table*}[t]
	\centering
	\small
	\caption{Definition of structured attributes in ``T3''. Rule-based attributes are computed from image-mask statistics, while semantic attributes are inferred from ``T2'' descriptions within constrained candidate sets. Percentages indicate the empirical distribution across all 7,778 images.}
	\label{tab:attribute_rules}
	\resizebox{\textwidth}{!}{
		\begin{tabular}{lll}
			\toprule
			\textbf{Type} & \textbf{Attribute} & \textbf{Candidate set (ratio)} \\
			\midrule
			\multirow{5}{*}{Rule-based}
			& Scale & tiny (5.3\%); small (26.1\%); medium (38.5\%); large (20.2\%); extensive (9.9\%) \\
			& Polarity & dark (79.6\%); neutral (9.8\%); bright (10.6\%) \\
			& Saliency & low (24.6\%); medium (50.2\%); high (25.2\%) \\
			& Number of defects & one (36.1\%); two (12.4\%); three (7.5\%); four (5.0\%); five (3.5\%); multiple ($>5$, 35.6\%) \\
			& Position & 3$\times$3 grid; defect-area ratio within each cell \\
			\midrule
			\multirow{3}{*}{Semantic}
			& Shape & \{linear, curved, elongated, circular, irregular, fragmented, network-like, spot-like, patch-like, diffuse, unclear\} \\
			& Direction & \{horizontal, vertical, diagonal, none\} \\
			& Spatial distribution & \{isolated, clustered, scattered, unclear\}; If the component count equals 1, forced as isolated. \\
			\midrule
			& Defect type & 25 predefined categories \\
			\bottomrule
	\end{tabular}}
\end{table*}

Table~\ref{tab:attribute_rules} summarizes the attribute definitions and discretization rules, along with the empirical distribution of each category across all 7,778 images. Fig.~\ref{fig:attribute_distribution_overview} further illustrates the distributions of key statistical attributes. The results show that the chosen discretization thresholds align well with the natural distribution of the data. For example, most samples fall within the small-to-medium scale range. Polarity is dominated by dark defects, while saliency exhibits a balanced distribution across low, medium, and high levels. Taken together, these observations indicate that the proposed attribute design is both data-driven and reasonably calibrated, providing a stable and interpretable intermediate representation for subsequent vision-language modeling.

\subsection{Social Impacts}
\label{subsec:Social_Impacts}
SteelDefectX is expected to contribute positively to industrial inspection by enabling more accurate and interpretable vision-language systems. Through its fine-grained, multi-form textual annotations, the dataset facilitates more effective human-model interaction by making predictions easier to understand, verify, and audit in real-world manufacturing scenarios. This can help reduce reliance on manual inspection and enhance product quality control. SteelDefectX is designed as a benchmark for systematic evaluation. Its use is intended to assist, rather than replace, human expertise, thereby supporting the development of reliable and accountable industrial inspection systems.

\subsection{Limitations}
\label{subsec:Limitations}
Despite its contributions, SteelDefectX has several limitations. First, natural language descriptions, by their nature, cannot fully capture the richness and subtlety of visual information. Nevertheless, language serves as an effective intermediate representation for structured communication, semantic abstraction, and alignment between visual and textual modalities. Second, some attributes (e.g., shape and direction) in ``T3'' rely on constrained semantic inference (see Table~\ref{tab:attribute_rules}), which may introduce minor inconsistencies. Third, while pixel-level segmentation masks are provided for all images, a portion of them are generated through semi-automatic pipelines and may still contain annotation noise, particularly for complex or low-contrast defects. Finally, SteelDefectX focuses specifically on steel surface defects. Future work may extend the annotation framework to broader industrial scenarios and further improve annotation consistency and coverage.

\section{Experiments}
\label{sec:A_exper}
\subsection{Experiments Compute Resources}
\label{subsec:Exper_compute_resources}
The implementation is based on PyTorch (v2.8.0), and all experiments are conducted on a local multi-GPU Linux server equipped with two Intel Xeon Silver 4314 CPUs, 188GiB system memory, and six NVIDIA GPUs. All reported training and evaluation jobs can be performed on a single NVIDIA RTX 4090 (24GB) GPU. The benchmark dataset contains 7,778 images resized to $256\times256$, resulting in modest storage and I/O requirements.

\subsection{Attribute-Level Localization Ablation}
\label{subsec:sup_attr}
We study how individual structured attributes affect dense visual grounding using FG-CLIP (ViT-L/14)~\cite{xie2025fgclip}. As shown in Table~\ref{tab:supp_fgclip_attr_loc_ablation}, adding a single attribute leads to different effects across metrics. \textit{DT+Shape} achieves the best IoU among single-attribute prompts, indicating that geometric cues are particularly effective for spatial localization. In contrast, attributes such as \textit{Position} and \textit{Spatial Distribution} do not improve pixel-level performance and even degrade AUROC compared with DT, despite yielding higher image-text similarity. Among full-text variants, ``T2'' achieves the best AUROC and F1-max, while ``T3'' provides slightly better IoU. ``T4'' attains the highest image-text similarity but shows weaker localization performance across all metrics. These results suggest that different attributes contribute differently to localization and classification. Different annotation forms (``T2''-``T4'') provide complementary strengths for vision-language learning, thus should be selected according to the requirements of specific tasks.

\begin{table*}[h]
	\centering
	\caption{Attribute-level localization ablation with frozen FG-CLIP (ViT-L/14) on SteelDefectX. ``DT'' denotes the defect-type-only prompt, while ``DT+attribute'' refers to adding a single structured attribute to the defect-type prompt. Best results are in bold.}
	\label{tab:supp_fgclip_attr_loc_ablation}
	\resizebox{0.73\linewidth}{!}{
		\begin{tabular}{lcccc}
			\toprule
			Text & AUROC & F1-max & IoU & ImgTxtSim  \\
			\midrule
			DT & 52.71 & 18.20 & 9.89 & 0.1719  \\
			\midrule
			DT+Shape & 58.29 & 20.13 & \textbf{11.02} & 0.1685  \\
			DT+Direction & 53.81 & 18.22 & 9.75 & 0.1857  \\
			DT+Spatial Distribution & 52.27 & 18.04 & 9.92 & 0.1971  \\
			DT+Number of Defects & 54.02 & 18.49 & 9.84 & 0.1662  \\
			DT+Position & 49.59 & 18.04 & 9.92 & 0.1872  \\
			DT+Scale & 54.22 & 18.44 & 9.81 & 0.1767  \\
			DT+Polarity & 53.85 & 18.23 & 9.91 & 0.1776  \\
			DT+Saliency & 55.88 & 19.08 & 9.85 & 0.1852  \\
			\midrule
			T2 & \textbf{58.54} & \textbf{20.48} & 10.03 & 0.1843  \\
			T3 & 58.07 & 20.06 & 10.14 & 0.1897  \\
			T4 & 54.53 & 18.43 & 9.45 & \textbf{0.1997}  \\
			\bottomrule
	\end{tabular}}
\end{table*}

\subsection{Additional Results}
\label{subsec:Additional_Results}
The experiments in this paper are conducted under largely deterministic settings, leading to minimal variability across runs. To assess stability, we perform a statistical analysis on representative configurations using Long-CLIP-Adapter with both ViT-B/16 and ViT-L/14 backbones. Each configuration is evaluated over three independent runs with different random seeds, and results are reported as mean $\pm$ standard deviation. The observed variance is consistently low (typically below 0.01 across metrics), and the relative performance ranking across text formats remains stable under both ``T0'' and ``T1''. Since the performance gaps across text formats are consistently larger than the observed standard deviations, the reported trends are considered stable and reliable. Therefore, we report single-run results in the main text for simplicity.

\begin{table*}[t]
	\centering
	\small
	\caption{Statistical significance analysis ($mean \pm std$ over 3 runs) using Long-CLIP-Adapter with different backbones.}
	\label{tab:statistical_results}
	\begin{tabular}{lcccc}
		\toprule
		\textbf{Setting} & \multicolumn{2}{c}{\textbf{ViT-B/16}} & \multicolumn{2}{c}{\textbf{ViT-L/14}} \\
		\cmidrule(lr){2-3} \cmidrule(lr){4-5}
		& \textbf{Acc} & \textbf{mAcc} & \textbf{Acc} & \textbf{mAcc} \\
		\midrule
		T2-T0 & $0.8759 \pm 0.0068$ & $0.8605 \pm 0.0121$ & $0.9227 \pm 0.0084$ & $0.9215 \pm 0.0093$ \\
		T2-T1 & $0.8279 \pm 0.0141$ & $0.7963 \pm 0.0124$ & $0.9096 \pm 0.0074$ & $0.8729 \pm 0.0057$ \\
		\midrule
		T3-T0 & $0.9002 \pm 0.0058$ & $0.8967 \pm 0.0043$ & $0.9314 \pm 0.0123$ & $0.9309 \pm 0.0048$ \\
		T3-T1 & $0.8411 \pm 0.0037$ & $0.8375 \pm 0.0069$ & $0.9316 \pm 0.0061$ & $0.9105 \pm 0.0113$ \\
		\midrule
		T4-T0 & $0.8931 \pm 0.0066$ & $0.8961 \pm 0.0073$ & $0.9234 \pm 0.0067$ & $0.9185 \pm 0.0031$ \\
		T4-T1 & $0.8150 \pm 0.0101$ & $0.8078 \pm 0.0039$ & $0.8957 \pm 0.0081$ & $0.8443 \pm 0.0099$ \\
		\bottomrule
	\end{tabular}
\end{table*}